\documentclass{ecai} 

\usepackage{latexsym}
\usepackage{amssymb}
\usepackage{amsmath}
\usepackage{amsthm}
\usepackage{booktabs}
\usepackage{enumitem}
\usepackage{graphicx}
\usepackage{color}

\usepackage{hyperref}

\usepackage{comment}
\usepackage{makecell}
\usepackage{pifont}
\usepackage{mathtools}
\usepackage{xspace}

\usepackage{algorithm}
\usepackage{algpseudocode}

\newcommand{\ie}{i.\,e.,\xspace}
\newcommand{\wrt}{w.\,r.\,t.\xspace}
\newcommand{\etal}{et~al.\xspace}

\newcommand{\BibTeX}{B\kern-.05em{\sc i\kern-.025em b}\kern-.08em\TeX}

\newcommand{\mymodel}{RA\-DAr\xspace}
\newcommand{\mymodellong}{Transformer-based Autoregressive Decoder Architecture}

\begin{document}


\begin{frontmatter}


\paperid{295} 


\title{\mymodel: A Transformer-Based Hierarchical Text Classifier with Autoregressive Decoder}
\title{\mymodel: A \mymodellong{} for Hierarchical Text Classification}


\author[A,D]{\fnms{Yousef}~\snm{Younes}\orcid{0000-0003-1271-3633}
\thanks{Corresponding Author. Email: yousef.younes@gesis.org}}
\author[B,C]{\fnms{Lukas}~\snm{Galke}\orcid{0000-0001-6124-1092}}
\author[D]{\fnms{Ansgar}~\snm{Scherp}\orcid{0000-0002-2653-9245}}

\address[A]{GESIS -- Leibniz-Institute for the Social Sciences }
\address[B]{Max Planck Institute for Psycholiguistics}
\address[C]{University of Southern Denmark}
\address[D]{University of Ulm}


\begin{abstract}
Recent approaches in hierarchical text classification (HTC) rely on the capabilities of a pre-trained transformer model and exploit the label semantics and a graph encoder for the label hierarchy. 
In this paper, we introduce an effective hierarchical text classifier \mymodel (\mymodellong) that is based only on an off-the-shelf RoBERTa transformer to process the input and a custom autoregressive decoder with two decoder layers for generating the classification output.
Thus, unlike existing approaches for HTC, the encoder of \mymodel has no explicit encoding of the label hierarchy and the decoder solely relies on the samples' label sequences observed during training.
We demonstrate on three benchmark datasets that \mymodel achieves results competitive to the state of the art with less training and inference time.
Our model consistently performs better when organizing the label sequences from children to parents versus the inverse, as done in existing HTC approaches.
Our experiments show that neither the label semantics nor an explicit graph encoder for the hierarchy is needed.
This has strong practical implications for HTC as the architecture has fewer requirements and provides a speed-up by a factor of 2 at inference time.
Moreover, training a separate decoder from scratch in conjunction with fine-tuning the encoder allows future researchers and practitioners to exchange the encoder part as new models arise.
The source code is available at \url{https://github.com/yousef-younes/RADAr}.

\end{abstract}

\end{frontmatter}


\section{Introduction}
\label{sec:intro}
Hierarchical Text Classification (HTC) deals with the task of labeling a text sample based on a semantic hierarchy where labels, \ie classes, exhibit a generalization-specialization relationship~\citep{li2020survey,silla2011survey,sun2001hierarchical,sebastiani2002machine}. In contrast to multi-label text classification, where the labels are treated as a flat set~\cite{galke2023really}, the models in HTC should exploit not only the text samples but also learn the labels' hierarchy during training.

Formally, in HTC, the set of labels $H$ is organized according to a label hierarchy where the labels are connected with a parent-child relationship~\cite{sebastiani2002machine}.
Each label represents a topical category, with the semantics of the relationship being that parent labels represent a broader category while child labels represent more specific categories.
When the hierarchy $\mathrm{H}$ is organized as a tree, each label $\mathrm{c\in H}$, except the root label, has one and only one parent. 
HTC approaches aim to predict a subset $\mathrm{L \subseteq H}$ for a given text instance. 
This subset contains one or more labels relevant to the text, taking the hierarchy into account. 
 
%
HTC approaches can be categorized into global and local methods based on how they utilize hierarchical information during training and how many classifiers they employ.
Local approaches employ one classifier per label or per level~\cite{banerjee2019hierarchical,shimura-etal-2018-hft}, whereas global approaches introduce a single classifier for all labels. Global approaches generally demonstrate superior performance compared to local methods~\cite{zhou-etal-2020-hierarchy}.
Recent approaches in HTC often utilize a pre-trained transformer model as the foundation of their architecture~\cite{jiang-etal-2022-exploiting,wang-etal-2022-incorporating, yu2022constrained}.

Global HTC models such as~\cite{zhou-etal-2020-hierarchy, deng2021htcinfomax, chen-etal-2021-hierarchy} 
often model the label hierarchy and text separately, and then find a mixed representation for the text and labels. To do that, each text sample interacts with the entire label hierarchy, leading to unnecessary calculations~\cite{zhou-etal-2020-hierarchy, chen-etal-2021-hierarchy}. HGCLR~\cite{wang-etal-2022-incorporating} partially solves this problem by using Graphormer~\cite{NEURIPS2021_f1c15925} to encode the hierarchy with the text and produce a hierarchy-aware text representation, which makes the Graphormer dispensable during testing~\cite{wang-etal-2022-incorporating}. A recent global model called HBGL~\cite{jiang-etal-2022-exploiting} achieves state-of-the-art results by integrating hierarchical information into BERT. It first pre-trains label embeddings on random paths sampled from the hierarchy and subsequently employs conditional generation to predict the labels, given the text.




In addition, Large Language Models (LLMs) such as GPT-4~\cite{achiam2023gpt} have recently been used to tackle the HTC task modeled as a multiple choice question~\cite{de2024ai}. The study shows that even for a very small hierarchy of 17 labels, the model was not able to achieve more than 50\% accuracy. Nevertheless, the literature on text classification shows that LLMs are strong text classifiers~\cite{DBLP:journals/corr/abs-2402-07470-pushing-the-limit,
sun2023text,
yuan2023revisiting,
li2023AreCA,
yu2023open}, especially given only a few in-context learning examples, but this performance comes with the extreme costs of dealing with large numbers of parameters while not generally outperforming smaller models. For instance, Yu \etal~\cite{yu2023open} confirm that by comparing the performance of the decoder-only models GPT-4~\cite{achiam2023gpt} and Llama 2~\cite{touvron2023llama} with the encoder-only RoBERTa~\cite{liu_roberta_2019} on three classification tasks and concluded that encoder-only models are better. 


The success of HTC models is largely attributed to incorporating linguistic knowledge found in the label names (label semantics) and explicitly encoding the label hierarchy using a graph encoder~\cite{chen-etal-2021-hierarchy, jiang-etal-2022-exploiting}. 
For instance, popular HTC approaches such as HiAGM~\cite{zhou-etal-2020-hierarchy}, HiMatch~\cite{chen-etal-2021-hierarchy}, and HGCLR~\cite{wang-etal-2022-incorporating} employ graph neural networks to process the hierarchy, while HBGL uses the adjacency matrix of the hierarchy to constrain the attention mechanism of transformer layers. 
 We question the necessity of such features and 
propose an effective sequence-to-sequence model \mymodel (\mymodellong{}) based on an off-the-shelf pre-trained transformer as an encoder and a comparably simple autoregressive decoder.
Thus, there is no graph encoder for the label hierarchy.
Our decoder learns the sequence of labels organized according to the hierarchy based on the samples.
Since symbolic labels are used instead of the original text of the labels, they provide no additional knowledge to the trained models~\cite{sebastiani2002machine}.
Thus, we do not exploit label semantics.
%
Our main contributions are: 
\begin{itemize}
    \item We propose \mymodel for hierarchical text classification.
    We use RoBERTa as an encoder and a custom autoregressive decoder with two decoder layers. 
    Unlike existing approaches, \mymodel does not need a graph encoder for the label hierarchy nor rely on label semantics.
    \item Experimental results on three HTC datasets confirm the effectiveness of the \mymodel in obtaining results competitive to the state of the art with considerably less training and half inference time.
    \item Unlike existing HTC approaches, which sort the labels from parents to children, we show that the inverse order from children to parents consistently produces better results.
\end{itemize}

Below, we summarize the related work.
Section~\ref{sec:method} introduces the \mymodel model. The experimental apparatus is described in Section~\ref{sec:experimentalapparatus}. Section~\ref{sec:results} reports an overview of the results. Section~\ref{sec:discussion} discusses the results and analyzes the model, before we conclude.
 \begin{table*}[h]
   \caption{Model comparison. Graph encoder is the component the model uses to encode the label hierarchy. Label order refers to the order the model receives the labels. Label semantics indicates using the linguistic knowledge found in the label names. CE stands for Cross-Entropy}
    \vspace{3mm}
    \centering

    \begin{tabular}{l|c|l|l|c|l|l}
        \toprule
         Model &\makecell{Label\\Hierarchy}&\makecell{Graph\\Encoder}&\makecell[lt]{Label Order} &\makecell{Label\\Semantics}&\makecell[lt]{Loss Function}&\makecell[lt]{Description}\\
        \midrule
            \multicolumn{7}{c}{Encoder-Only Baselines}  \\
        \midrule

        BERT&No&No&No&No&CE&Transformer Model\\
        RoBERTa&No&No&No&No&CE&Transformer Model\\
        
        \midrule
            \multicolumn{7}{c}{Encoder-Decoder Baselines}  \\
        \midrule
        BART&No&No&No&Yes&CE&Transformer-based, denoising autoencoder\\
        T5 &No&No&No&Yes&CE&Transformer-based, text-to-text framework\\

        \midrule
            \multicolumn{7}{c}{Hierarchical Text Classifiers (based on pre-trained encoder-only models)}  \\
        \midrule
        HiAGM &Yes&\makecell[lt]{Tree-LSTM,\\GCN}&Parent-Child&Yes&\makecell[tl]{BCE and \\Recursive Regularization}&\makecell[tl]{Obtains label-wise text features by fusing text\\ classification model with a hierarchy encoder}\\
        HTCInfoMax &Yes&\makecell[lt]{Tree-LSTM,\\GCN}&Parent-Child&Yes&\makecell[tl]{Text-label MIM,\\Label prior matching, and BCE.}&\makecell[tl]{Uses MIM to improve text-\\label association and label representation.}\\
        HiMatch &Yes&GCN&Parent-Child&Yes&\makecell[lt]{Joint embedding, Hierarchy-\\aware matching, and CE}&\makecell[tl]{Models the text-label semantics\\ relation as a semantic matching problem.}\\
        HGCLR &Yes&\makecell[lt]{Modified \\Graphormer}&Parent-Child&Yes&\makecell[lt]{NT-Xent and  BCE}&\makecell[lt]{Uses contrastive learning to produce\\ hierarchy-ware text representation}\\
        HBGL &Yes&BERT&Parent-Child&Yes&\makecell[tl]{BCE}&\makecell[lt]{Uses Attention Mask in self-attention\\ layers to feed BERT with label graph} \\

        \midrule
        \multicolumn{7}{c}{Hierarchical Encoder-Decoder Models}  \\
        \midrule
        SGM&Yes&No&Parent-Child&Yes&CE&\makecell[tl]{Uses label correlations and text-\\label relations}\\
        Seq2Tree&Yes&No&Parent-Child&Yes&CE&\makecell[tl]{Uses T5 with the decoder\\ constrained by the hierarchy}\\
        
        \midrule
        
        \mymodel &Yes&No&Child-Parent&No&\makecell[lt]{Focal Cross Entropy}&\makecell[tl]{Uses RoBERTa as encoder and \\an autoregressive decoder on symbolic labels
        }\\
       
         \bottomrule
    \end{tabular}
    \label{tab:model_comparison}
\end{table*}

\section{Related Work}
\label{sec:literature}

There are two types of HTC models, namely local and global ones~\cite{zhou-etal-2020-hierarchy}. Recently, the focus has been on global models thanks to the advent of word embeddings and pre-trained language models, which encourage general solutions. We focus on global methods, as the best-performing HTC models like~\cite{jiang-etal-2022-exploiting, wang-etal-2022-incorporating} belong to this type.

We present a set of models that were shown to be effective for HTC, irrespective of their initial design intent. 
These models are grouped into encoder-only and encoder-decoder based on their architecture, with chronological ordering within each category.
We compare the models regarding five distinguishing features shown in Table~\ref{tab:model_comparison}.
The features are the (i)~use of the label hierarchy provided by the dataset (or considering the labels as set), 
(ii)~employment of a graph encoder to represent the hierarchy, 
(iii)~information on the order in which the model receives the labels, which is either parents to children or vice versa,
(iv)~use of label semantics, \ie the textual description of the label,
and 
(v)~the used loss function. 

\paragraph{Encoder-only Models} 
Encoder-only transformer models like BERT~\cite{devlin_bert_2019} and RoBERTa~\cite{liu_roberta_2019} have shown superior classification performance in supervised settings~\cite{yu2023open}. 
They perform well on the HTC task even without considering the label hierarchy~\cite{galke2023really}.

BERT performs the classification task by encoding the labels as a multi-hot vector and feeding them along with their associated texts into the model, which uses a classification head to do the prediction~\cite{devlin_bert_2019}. 
By this, BERT ignores the hierarchy \ie treats the labels as a set like in a multi-label task. 

RoBERTa is an optimized version of BERT that drops the next sentence objective and uses more data for training~\cite{liu_roberta_2019}.
Due to their strong performance, the encoder-only models are used as a foundation by many HTC methods.

Another strong approach is HiAGM, which uses two encoders, one for text and one for encoding the label hierarchy~\cite{zhou-etal-2020-hierarchy}. 
The model provides two variants for the hierarchy encoder: a Tree-LSTM and GCN.
The label dependencies are modeled bidirectionally, \ie from parents to children and vice versa. As the loss function, this model combines recursive regularization~\cite {gopal2013recursive} for the parameters of the final fully connected layer with binary cross-entropy (BCE).

HTCInfoMax builds on a variant of the HiAGM model with a modified loss function, which combines text-label Mutual Information Maximization (MIM), label prior matching, and BCE. 
The MIM helps to improve the text-label association. 
HTCInfoMax addresses the inherent label imbalance issue in HTC by considering statistical constraints on the structural encoder to improve the representation of low-frequency labels~\cite{deng2021htcinfomax}.

HiMatch uses GCN as its graph encoder~\cite{chen-etal-2021-hierarchy}. It addresses the text-label association in terms of semantic matching. The system utilizes a three-component loss function. The joint embedding loss aligns text and label semantics, while the hierarchy-aware matching loss emphasizes semantic proximity of child labels to text semantics over parent labels. In addition, cross-entropy loss is used to learn the relationship between the text and the labels. 

HGCLR~\cite{wang-etal-2022-incorporating} incorporates the hierarchy using a modified Graphormer into a BERT representation of the text~\cite{NEURIPS2021_f1c15925}.
It uses the label hierarchy to construct positive samples for contrastive learning.
The NT-Xent loss~\cite{chen2020simple} is applied to learn hierarchy-aware text representations by pulling together the text input close to its positive samples. 
In prediction, the model only needs the text encoder, not the Graphormer, so it becomes a BERT-encoder with classification head. 

The state-of-the-art hierarchical text classifier, HBGL, utilizes the encoder-only BERT model in two steps. Initially, it pre-trains label vectors based on the hierarchy. Subsequently, the BERT model and these pre-trained label vectors are fine-tuned to predict the corresponding labels. This prediction occurs level-wise by constraining the attention matrix~\cite{jiang-etal-2022-exploiting}. The training process employs BCE loss. 


\paragraph{Encoder-Decoder Models} 
An encoder-decoder model approaches the HTC task as a text generation problem. It comprises an encoder for processing the text input and a decoder for generating the corresponding labels.

SGM is a Bi-LSTM encoder-decoder model with attention mechanism~\cite{yang-etal-2018-sgm}. 
The encoder uses the attention mechanism to produce a context vector focusing on informative words. The decoder uses the context vector, hidden state of the previous step, and label embedding vector to generate the hidden state of the current step. SGM has no graph encoder but benefits from label semantics and uses cross-entropy loss. 
    
BART is a transformer-based 
encoder-decoder with an encoder similar to BERT and an autoregressive decoder similar to GPT~\cite{radford2018improving}. Even though it does not have a graph encoder, we use it for HTC by framing the task as text generation since BART is effective when fine-tuned for text generation~\cite{lewis-etal-2020-bart}. The model generates the labels in whatever order we use during fine-tuning but utilizes label semantics. It is trained using cross-entropy loss.

T5 is also a transformer-based encoder-decoder model that supports various NLP tasks in a text-to-text format with a different prefix for every task~\cite{raffel2020exploring}. Like BART, T5 frames the HTC task as text generation, benefits from label semantics, and uses cross-entropy loss.  

Similar to BERT and RoBERTa, the encoder-decoder models BART and T5 are not designed for HTC tasks.
They do not provide an explicit graph encoder.
Thus, for the hierarchical multi-label classification task, the labels are treated as a set.
We include these models as baselines since they have shown strong performance despite not considering the hierarchy information~\cite{galke2023really}.

An approach considering the label hierarchy is Seq2Tree~\cite{yu2022constrained}.
It is a variant of the T5-base model and captures the hierarchical information using the Depth-First Search (DFS)~\cite{doi:10.1137/0201010}
over the hierarchy to linearize the labels.
The decoder uses the encoder output to generate the DFS label sequence one at a time, considering the hierarchical information. 
In other words, the candidate labels for one step are the children of the parent predicted in the previous step. 

\section{The \mymodel Model}
\label{sec:method}
\begin{figure}[htbp]
  \centering  
  \includegraphics[scale=0.8]{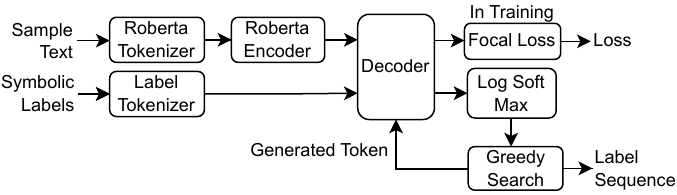} 
  \vspace{-4mm}
  \caption{The \mymodel Model Architecture}
  \label{fig:model}
\end{figure}

In this section, we introduce the \mymodel model depicted in Figure~\ref{fig:model}. \mymodel is a transformer-based encoder-decoder model. 
Any encoder-only model can be used as the encoder of \mymodel, but we choose to use RoBERTa base. The decoder is an autoregressive transformer-based model trained from scratch to generate a sequence of labels.
The output of the next label is conditioned on the previous labels to generate the labels along the hierarchy for the HTC task. 
The model employs a modified version of focal loss~\cite{lin2017focal} to pay more attention to difficult samples with high loss. 
In the following, we briefly describe the use of RoBERTa as an encoder.
Then, we explain the decoder and the training and testing procedures. 

\begin{figure*}[ht]
  \centering
  \includegraphics[width=\textwidth]{images/label_preprocessing.pdf}
  \vspace{-4mm}
  \caption{Label preprocessing and tokenization. Line a) contains the original labels. Line b) maps the original labels to the symbolic labels. Line c) adds the level separator token <unk>. Line d) organizes the labels level-wise from children to parents. Line e) contains the padded tokenizer output.}
   \vspace{1mm}
   \label{fig:tokenization}
\end{figure*}

\paragraph{Encoder}
\label{encoder}
The model uses the RoBERTa base as its encoder part. The encoder is responsible for producing a fixed-size context tensor that captures the linguistic information found in the input text. To obtain the context tensor, the text is fed into the RoBERTa tokenizer to produce the token indices and attention masks. These are then fed into the RoBERTa model to produce an output. We use the last hidden state of that output, a tensor of size $(512,768)$, as input to the decoder.
For the decoder, we expand the dimensions of the attention masks to avoid attending to padding tokens.

\paragraph{Decoder}
\label{sec:decoder}
The decoder is an autoregressive transformer model whose vocabulary corresponds to the label set. As such, the decoder's vocabulary is limited to the symbolic labels~\cite{sebastiani2002machine}, \ie internal label identifiers and special tokens (see Figure~\ref{fig:tokenization}). Our decoder iteratively uses the previously generated tokens and the encoder's output as condition to generate the next token. It uses greedy search that picks the token with the highest probability at each decoding step without considering future consequences~\cite{wilt2010comparison}.
It stops when it reaches the special token \texttt{</s>}, indicating the end of the sequence.

In more detail, the decoder consists of embedding and position embedding layers, followed by decoder layers and a linear layer to produce the logits. The whole procedure of the decoder is shown in Algorithm 1. It receives four inputs during training: the sample's labels $\mathrm{L \subseteq  H}$, the labels' mask $\mathrm{l_m}$, the last hidden state of the encoder output $\mathrm{h}$, and the encoder mask $\mathrm{m}$. It starts by producing the absolute position embeddings $\mathrm{l_e}$ for the labels to capture the hierarchical information from the ordered label sequence. Then, it passes the obtained label embeddings $\mathrm{l_e}$ along with $\mathrm{h}$, $\mathrm{m}$, and $\mathrm{l_m}$ through a decoder layer which performs three operations.
First, it computes the multi-head self-attention $\mathrm{att}$ of the label embedding. Next, it passes the obtained $\mathrm{att}$ through a 
layer normalization~\cite{ba2016layer} and dropout~\cite{hinton2012improving} layers to compute the query $\mathrm{q}$.
After that, the encoder's last hidden state $\mathrm{h}$ is used as the key and value and sent together with the obtained query and the encoder mask $\mathrm{m}$ to a transformer block to perform cross attention~\cite{NIPS2017_3f5ee243} and produce the output of the decoder layer. The encoder attention mask $\mathrm{m}$ controls the decoder attention on the encoder hidden states.

The output of a decoder layer is used as the label embedding for the next decoder layer that repeats the same operations.
Furthermore, a linear layer is used to produce the logits on the output of the last decoder layer. Finally, the logits are used to compute the focal loss, a scaled cross-entropy that pays more attention to difficult samples.

\begin{algorithm}
\caption{Forward Pass of \mymodel Decoder}
\begin{algorithmic}[1]
    \State \textbf{Input:} \parbox[t]{\dimexpr\linewidth-\algorithmicindent}
    {
       $\text{labels}\ L, \text{label\ masks}\ l_m$ \\
       $\text{encoder\ output}\ h\text{, encoder\ mask}\ m$
    }
    \State \textbf{Output:} \text{loss}
    \State $l_e \gets \operatorname{Dropout}(\operatorname{word-embed}(L)+\operatorname{pos-embed}(L))$

    \For{$i = 1$ \textbf{to} $N_\mathrm{layers}$}
        \State $\text{att} \gets \operatorname{MultiheadSelfAttention_i}(l_e,l_e,l_e,l_m)$
        \State $q \gets \operatorname{Dropout}(\operatorname{LayerNorm_i}(\text{att} + l_e))$
        \State $l_e \gets \operatorname{TransformerBlock_i}(h, h, q, m)$
    \EndFor
    \State $\text{logits} \gets \operatorname{linear}(l_e)$
    \State $\text{loss} \gets \operatorname{cross-entropy}(L,\text{logits})$
    \State $\text{loss} \gets (1 - e^{-\text{loss}})^{\gamma} \cdot \text{loss}$
    \Comment{focal loss adjustments}
    \State \textbf{return} $\text{loss}$
\end{algorithmic}
\end{algorithm}

Using the WOS dataset, we conducted pre-experiments regarding the number of decoder layers and the features of its transformer blocks~\cite{NIPS2017_3f5ee243}. We observe that the optimal choice is when the decoder has two decoder layers. Each layer has a transformer block with eight attention heads, a feedforward network and uses a dropout of 0.2. 
\paragraph{Training and Testing}
The labels in HTC datasets are typically very imbalanced~\cite{DBLP:journals/ir/VaglianoGS22}.
To compensate for this imbalance, we employ a batch-level focal loss~\cite{lin2017focal}, \ie the model aims to optimize batches that are hard to classify. Those usually include samples with low-frequent labels in the training data.

We optimize the model to minimize cross-entropy on the batch level with the ground truth and apply focal loss scaling (line 11 in Algorithm 1). To achieve that, we calculate the model confidence on the batch $e^\mathrm{-loss}$. Then, we compute the model perplexity $1-e^\mathrm{-loss}$. After that, we raise the perplexity to the power of the focusing parameter $\gamma$ to get the modulating factor. We use the modulating factor to adjust the batch cross-entropy loss based on the batch difficulty. If the loss is high, the model confidence is low, and its perplexity is high, resulting in a higher final loss. On the contrary, when the loss is low, the confidence is high, and the perplexity is low, resulting in a lower final loss\label{sec:train}.


During testing, the decoder starts with the beginning-of-sequence token \texttt{<s>}, generates tokens autoregressively, one at a time, and stops generating labels after reaching the end-of-sequence token \texttt{</s>}. \mymodel uses greedy search to generate the tokens, and using beam search~\cite{wilt2010comparison} does not improve the results. When the model works on a batch of samples, the \texttt{</s>} token is predicted for each sample independently.

\section{Experimental Apparatus}
\label{sec:experimentalapparatus}

\begin{table*}[ht]
\caption{Statistics of the datasets. 
$|H|$ is the number of Labels. $D$ is the maximum level of hierarchy.
$\mathrm{Avg}(L_i)$ is the average number of labels per sample. $\mathrm{Avg}(PL_i)$ is the average number of parent labels per sample. $\mathrm{Avg}(LL_i)$ is the average number of leaf labels per sample. These average fields show the statistics for train, dev, and test sets.}
 \vspace{3mm}

    \centering
    \begin{tabular}{c|cccccccc}
    \toprule
        Dataset & $|H|$ & $D$ & $\mathrm{Avg}(L_i)$& $\mathrm{Avg}(PL_i)$ &$\mathrm{Avg}(LL_i)$ & Train & Dev & Test \\
    \midrule
         WOS & 141 & 2 & 2.0/2.0/2.0& 1.0/1.0/1.0 &1.0/1.0/1.0&30,070 & 7,518 & 9,397\\
         NYT& 166 & 8 & 7.6/7.6/7.5& 5.9/5.9/5.9 &1.7/1.6/1.7&23,345 & 5,834 & 7,292\\
         RCV1-V2 & 103 & 4 &3.18/3.18/3.24&1.98/1.98/2.02&1.20/1.21/1.23& 20,833 & 2,316 & 781,265 \\
    \bottomrule
    \end{tabular}
    \label{tab:datasets}
\end{table*}

\paragraph{Datasets and Metrics}
\label{sec:datasets_measures}
We use three common HTC datasets. Those are Web-of-Science (WOS) ~\cite{kowsari2017hdltex}, NY-Times (NYT)~\cite{shimura-etal-2018-hft}, and RCV1-V2~\cite{lewis2004rcv1}. 
The WOS dataset has a hierarchy of two levels. It has exactly two labels per sample, one child and its parent, which makes it easier than the other two datasets. 
The NYT and RCV1-V2 datasets have eight-level and four-level hierarchies, respectively. 
The RCV1-V2 dataset has the largest test set compared to the others. Table~\ref{tab:datasets} shows statistics about these datasets. 

We measure the experimental results using Micro-F1 and Macro-F1 metrics. The impact of rare labels is higher for Macro-F1 score than for Micro-F1.
For \mymodel experiments, we report the mean and standard deviation over five runs with different random seeds.


\paragraph{Data Preprocessing}
\label{sec:preprocessing}
The labels in the hierarchies of the datasets are usually provided with plain, human-readable text. 
Previous works \cite{wang-etal-2022-incorporating, jiang-etal-2022-exploiting} tried to benefit from the linguistic knowledge found in the human-readable label names, known as label semantics~\cite{sebastiani2002machine}. 
On the contrary, our model does not rely on such label semantics.

Rather, we replace the original labels with symbolic labels. For example, the symbolic label \texttt{[a\_14]}  represents the label ``Top/Classifieds/Job Market/Job Categories/Media, Entertainment and Publishing''  in the NYT dataset as shown in Figure~\ref{fig:tokenization}.

The label tokenizer's vocabulary consists of the set of symbolic labels of the dataset.  The label tokenizer is responsible for tokenizing the input symbolic labels and generating a fixed-size vector whose size accommodates the maximum possible number of labels (not the average) for the dataset and other special tokens. 
This vector size differs from one dataset to another; 
it is $6$, $48$, $22$ for the WOS, NYT, and RCV1-V2 datasets.

The tokenizer uses the \texttt{<s>} and \texttt{</s>} tokens to mark the beginning and end of the label sequence. 
It also uses the \texttt{<pad>}  token to fill the rest of the vector after the \texttt{</s>} token. In addition, the \texttt{<unk>} token separates hierarchy levels. 

The exact purpose of the \texttt{<unk>} token is demonstrated in an example shown in Figure~\ref{fig:tokenization}.
It shows the label preprocessing and tokenization on one label set corresponding to a sample from the NYT dataset. 
The line a) contains the original labels organized hierarchy level-wise from parents to children. 
Line b) contains the symbolic labels corresponding to the original labels. 
Line c) shows how the \texttt{<unk>} token separates the labels from different levels. 
Line d) represents the tokenizer's input and contains the labels organized level-wise from children to parents. 
Line e) expresses the tokenizer's output given line d) as input.  

\paragraph{Procedure and Hyperparameters}
\label{sec:procedure}
During training, we use the AdamW optimizer~\cite{loshchilov2017decoupled} with default settings and learning rates of $5\cdot10^{-5}$ and $3\cdot10^{-4}$ for the encoder and decoder, respectively. Two learning rates are necessary because the encoder is pre-trained while the decoder is trained from scratch. We also use the 
ReduceLROnPlateau scheduler~\cite{10.1007/978-3-031-08277-1_17} with patience of 3.
After each epoch, the average validation loss is used to measure progress. If the validation loss does not decrease for three epochs, the scheduler multiplies the learning rates by a factor of $0.1$ for both the encoder and decoder. The encoder's learning rate decreases until it is lower than $5\cdot10^{-7}$. 
At that point, the encoder is frozen, and the decoder continues to learn alone. The training loop has a patience of $10$, so it stops if the average validation loss does not decrease for ten consecutive epochs. We use focal loss with focusing parameter $\gamma=2$ as described in Section~\ref{sec:train}. Focal loss depends on cross-entropy with label smoothing of $0.1$.
We use teacher forcing~\cite{william1989} and gradient accumulation over $2$ batches ($32$ examples each). For reproducibility, we use a fixed random seed. With these settings, the model took $29$, $30$, and $40$ training epochs on the WOS, NYT, and RCV1-V2 datasets, respectively. 
With a patience of $10$, the model saturates at $23$ epochs on average.
The experiments were implemented in PyTorch and run on an NVIDIA A100-SXM4-40GB GPU. 

The hyperparameter values like learning rates, batch size, focusing parameter, and label smoothing are selected per dataset based on the validation sets using wandb~\cite{wandb} as follows. We specify ranges for the values depending on the common values from the literature and also based on our pre-experiments. For example, $5\cdot10^{-5}$ is known to be a reasonable learning rate for RoBERTa~\cite{galke2023really}, so we specify the range $[5\cdot10^{-4},5\cdot10^{-6}]$ to contain that value. 
Then, using these ranges, we initialize each hyperparameter randomly over $20$ experiments and select the best value. We use the best values for individual hyperparameters and test all their possible combinations, \ie grid search~\cite{wandb}, again on the validation sets. The values used to select the hyperparameters are in the code repository associated with the paper.
\begin{table*}
\centering
\caption{Hierarchical text classification results. The \mymodel model results are the average over five runs and are reported along with the standard deviation.  For numbers from the literature, "-" means not available in the original paper.}
\vspace{3mm}
\begin{tabular}{l|c|c|c|c|c|c|c}
\toprule
           Model    & \multicolumn{2}{c}{WOS} & \multicolumn{2}{c}{NYT}    & \multicolumn{2}{c}{RCV1-V2} & Provenance   \\ 
                        & Micro-F1           & Macro-F1        & Micro-F1           & Macro-F1 & Micro-F1           & Macro-F1 &     \\ \hline
\multicolumn{8}{c}{Encoder-Only Baselines}  \\
\midrule
BERT-base &85.63&79.07&78.24&65.62&85.65&67.02&\cite{wang-etal-2022-incorporating}\\
RoBERTa &-&-&77.05&55.53&86.99&62.29&\cite{galke2023really}\\
\midrule
\multicolumn{8}{c}{Encoder-Decoder Baselines}  \\
\midrule
BART &84.08&77.43&19.21&6.49&86.20&65.11&our experiment\\
T5   &82.03&74.62&46.71 & 20.06&84.9 &57.01 &our experiment\\
\midrule
\multicolumn{8}{c}{Hierarchical Encoder-Only Classifiers}  \\
\midrule
HGCLR & 87.11 & 81.20& 78.86&67.96&86.49&68.31&\cite{wang-etal-2022-incorporating}\\
BERT+HiAGM &86.04&80.19&78.64&66.76&85.58&67.93&\cite{wang-etal-2022-incorporating}\\
BERT+HTCInfoMax &86.30&79.97&78.75&67.31&85.53&67.09&\cite{wang-etal-2022-incorporating}\\
BERT+HiMatch &86.70&81.06&-&-&86.33&68.66&\cite{chen-etal-2021-hierarchy}\\
HBGL+RoBERTa  &87.66&81.96&79.98&70.57&87.52&70.52&our experiment\\
HBGL+BERT&\textbf{87.36}&82.00&\textbf{80.47}&70.19&\textbf{87.23}&\textbf{71.07}&\cite{jiang-etal-2022-exploiting}\\
\midrule
\multicolumn{8}{c}{Hierarchical Encoder-Decoder Models}  \\
\midrule
SGM &67.74&74.01 & 64.68&\textbf{72.78} & 71.85 & 35.29 & our experiment  \\
SGM-T5 &85.83&80.79&-&-&84.39&65.09&\cite{yu2022constrained}\\
Seq2Tree &87.20&\textbf{82.50}&-&-&86.88&70.01&\cite{yu2022constrained}\\
\midrule
\mymodel &$87.17_{(0.04)}$&$81.84_{(0.08)}$&$79.84_{(0.07)}$&$68.64_{(0.28)}$&$87.23_{(0.05)}$&$69.64_{(0.12)}$& our experiment \\
\textit{difference to HBGL}   & -0.19 & -0.16 & -0.63 & -1.55 & 0.0 & -1.43 & \\
\bottomrule
\end{tabular}
\label{tab:results}
\end{table*}
\section{Results}
\label{sec:results}

Table~\ref{tab:results} shows Micro-F1 and Macro-F1 on three datasets.
We compare our model to the models
from the literature (see Section~\ref{sec:literature}). 
Some hierarchical text classification models like HiMatch and HiAGM did originally not use pre-trained language models as text encoder. Yet, the HGCLR paper~\cite{wang-etal-2022-incorporating} shows that replacing the text encoder in such models with a BERT model leads to improved results, which we use for our comparison.

Table~\ref{tab:results} is
organized into five groups, depending on the models' architecture.
The first group reports the results for the encoder-only baseline models BERT~\cite{devlin_bert_2019} and RoBERTa~\cite{liu_roberta_2019}, which treat the labels as a flat set.
The second group includes encoder-decoder baseline models T5~\cite{raffel2020exploring} and BART~\cite{lewis-etal-2020-bart}, which yield lower performance than the encoder-only baselines. 
The third group comprises HTC models using encoder-only models, namely HiAGM~\cite{zhou-etal-2020-hierarchy}, HTCInfoMax~\cite{deng2021htcinfomax}, HiMatch~\cite{chen-etal-2021-hierarchy}, HGCLR~\cite{wang-etal-2022-incorporating}, and HBGL~\cite{jiang-etal-2022-exploiting}.
We report the results of HiAGM, HTCInfoMax, and HiMatch models with their original text encoder replaced with BERT. The highest scoring model in this group is  HBGL. 
We report the results of HBGL using both BERT and RoBERTa.

The fourth group reports the results of the following hierarchical encoder-decoder models SGM, SGM-T5, and Seq2Tree. 
SGM uses its own vocabulary and trains its word embedding from scratch~\cite{yang-etal-2018-sgm}. Nevertheless, SGM beats all other models on the Macro-F1 score of the NYT dataset.
%
SGM-T5 outperforms SGM. It is a modified version of SGM that uses T5 as its encoder-decoder backbone~\cite{yu2022constrained}.  
The Seq2Tree model is the best-performing model in this group. 
It even achieves a Macro-F1 score better than all other models on the WOS dataset. 
Because SGM and Seq2Tree models outperform HBGL only in isolated metrics across specific datasets, we regard HBGL as the current state-of-the-art.

The fifth group contains the results of our \mymodel model. 
It outperforms the encoder-only baselines by achieving improvements in Macro-F1 of 2.77\%, 3\%, and 2.62\% on WOS, NYT, and RCV1-V2, respectively.
Compared to HBGL, the \mymodel model achieves a similar Micro-F1 score on the RCV1-V2 dataset. For the other two datasets, it achieves results with differences ranging from $0\%$ to $1.55\%$ in all measures. Here, we should indicate that we compare \mymodel to HBGL with the BERT encoder since it is the one used in the original paper. However, using RoBERTa as HBGL's encoder does not make a big difference in the results. Nevertheless, including those results is important to highlight the competitiveness of \mymodel \wrt HBGL.
Finally, our \mymodel model outperforms most HTC models, while it does not use label semantics or a graph encoder for the hierarchy. 
\section{Discussion}
\label{sec:discussion}
The results show that RADAr is competitive with other HTC methods, providing a strong indication that an explicit graph encoder is not needed as long as the labels are ordered according to the hierarchy. This conceptual advantage has an immediate effect on training and inference times. Although \mymodel has more parameters than HBGL ($137$M vs. $110$M), it requires less time for training and inference. Averaged over our three datasets, HBGL needs $48$ epochs, each taking $11.67$ minutes, while \mymodel needs $33$ epochs, each taking $8.7$ minutes. Comparing inference times, we found that HBGL requires approximately $135$, $184$, and $12,406$ seconds compared to $74$, $82$, and $6,716$ on the WOS, NYT, and RCV1-V2 test sets, respectively. 
On average, \mymodel provides a speed-up of $97.03$\% for inference time relative to its strongest competitor HBGL -- effectively doubling the throughput for practical applications. 


The primary distinction of our \mymodel model from previous models is that it does not use a graph encoder or label semantics. Instead, \mymodel captures hierarchical information from the organization of the label sequences it is trained on. To understand the impact of label organization and the components of \mymodel, we conduct an ablation study (Section~\ref{subsec:ablation study}) and an error analysis  (Section~\ref{subsec:error_analysis}).

\begin{table*}[!t]
\centering
\caption{The ablation experiment's nomenclature reflects feature deviation from our RADAr model. RADAr features include using RoBERTa as an encoder and focal loss on the batch level. It
also include working on labels ordered from child-parent hierarchy level-wise using the <unk> token to separate labels from different levels and without using label semantics.}
 \vspace{3mm}
\label{tab:ablation}

\begin{tabular}{l|c|c|c|c|c|c}
\hline
           Experiment    & \multicolumn{2}{c}{WOS} & \multicolumn{2}{c}{NYT}    & \multicolumn{2}{c}{RCV1-V2}\\   

                        & Micro-F1           & Macro-F1        & Micro-F1           & Macro-F1 & Micro-F1           & Macro-F1\\
        
\hline
\mymodel model\ldots &87.19&81.84&79.88&\textbf{69.09}&87.22&69.48\\
\hline

\ldots but parent-to-child ordering &86.55&81.20&79.22&66.43&87.04&68.57\\
\ldots w/o <unk> token separators  &\textbf{87.20}&\textbf{81.85}&76.28&65.40&85.88&64.05\\
\ldots using <unk> to separate paths instead of levels &87.19&81.84&77.35&62.95&79.35&63.64\\
\ldots but shuffled labels w/o <unk> &86.16&80.77&67.08&47.50&86.12&64.30\\
\ldots children only + hierarchy &87.11&81.61&\textbf{79.94}&68.49&87.13&\textbf{69.61}\\
\hline
\ldots focal loss on label level &86.64&81.10&80.11&69.0&87.10&69.36\\
\ldots w/o focal loss &87.01&81.41&79.55&68.37&\textbf{87.25}&69.40\\
\hline
\ldots with labels semantics &86.58&81.49&79.60&67.52&86.96&69.51\\
\hline
\ldots BERT Encoder &86.51&80.77&78.93&67.19&87.05&68.64\\
\ldots XLNET-base Encoder &86.98&81.60&79.32&67.26& 86.60 & 68.15 \\
\ldots Sentence Transformer Encoder &86.76&81.15&76.77&63.89&84.68&62.31\\ 
\hline
\end{tabular}
\end{table*}

\subsection{Ablation Study}
\label{subsec:ablation study}
We analyze the different components of \mymodel to understand their impact. The ablation study covers different variants of label sequence organization, the effect of focal loss, the effect of label semantics, and employing different text encoders.
Table~\ref{tab:ablation} summarizes the results of our ablation studies, beginning with the best scores achieved with our \mymodel, which we describe in the following.

First, we compare the results of organizing the label sequence from parents to children versus children to parents (see also the example in Figure~\ref{fig:tokenization}).
Our ablation demonstrates that ordering the labels from parents to children decreases all measures over all datasets. This drop reaches its maximum  $2.66\%$ less Macro-F1 score on the NYT. Ordering the labels children to parent positively affects all measures. This effect relates to the depth of the hierarchy, \ie the deeper the hierarchy is, the more remarkable the effect is.  

The experiment without the use of the \texttt{<unk>} token as a separator (see line `` w/o <unk> '') shows a small negative impact on the WOS dataset but is essential for the NYT and RCV1-V2 datasets. 
In this experiment, the model runs on the labels organized level-wise from child to parent but does not use the \texttt{<unk>} token to indicate the different levels of the hierarchy. 
The small effect on WOS since is expected because WOS has exactly two labels: one parent and one child.
Adding the \texttt{<unk>} token is not very helpful in this case.

The \mymodel model uses \texttt{<unk>} as a level separator to delimit different hierarchy levels. Instead one could use the \texttt{<unk>} token to separate different paths of the same sample (see line ``using <unk> to separate paths instead of levels''). The model runs on the full paths, where a path is constructed starting from a child label up to the root. As the results show, organizing the labels in this way makes the problem more difficult for datasets with complex hierarchies. Again, it does not affect the WOS dataset, which only has a two-level hierarchy and only one assigned leaf label per document.

To show the influence of organizing the label sequence,
the ``shuffled labels w/o <unk>'' experiment fine-tunes the model on shuffled label sequences. 
The results show that organizing the labels is essential for the model to perform well. 
Since the model cannot pick up a random but fixed order of labels, it also shows that the order in terms of the child to parent actually is a simplified form of entailment where the child is the premise and its parent is the hypothesis. In other words, the model is doing a form of inference on the symbolic labels~\cite{maccartney2009natural}.
Again, the effect of organized labels depends on the depth of the label hierarchy in the dataset. For example, the Macro-F1 score drops by $1.07\%$, $5.18\%$, and $21.59\%$ for the WOS, RCV1-V2, and NYT datasets, respectively. 
This also indicates that the deeper the hierarchy is, the more effect the label organization has. 

Next, we simplify the HTC task using the label hierarchy since every child label has only one parent. If the child label is predicted, we can use the hierarchy to extract all its parents up to the root. The ``children only + hierarchy'' experiment runs the model on the datasets by removing all the labels that are extractable from the hierarchy based on other labels. 
For every sample, we start from the children labels in the sequence and include a label if it has no children in the sequence. Applying this method on line d)~of Figure~\ref{fig:tokenization} results in \texttt{[[a\_14], <unk>, [a\_37], <unk>, [a\_42], [a\_35], <unk>]}, which contains the minimum label set necessary to reconstruct line d) using the hierarchy. 
Such a scenario is feasible if the hierarchy is available at test time, which is commonly the case as these hierarchies are typically openly available~\cite{DBLP:conf/kcap/Grosse-BoltingN15,
DBLP:conf/jcdl/MaiGS18}.
After the model generates the minimized label set, we use the hierarchy to extract the remaining labels up to the root. The model performs well in this scenario and produces a slightly better Micro-F1 score on the NYT dataset than our best-performing model. This improvement is due to the problem's simplification, which is more noticeable given the complexity of the NYT dataset.


The third group of experiments in Table~\ref{tab:ablation} investigates the effect of focal loss. In the ``focal loss on label level'' experiment, the model runs with the focal loss computed on every label. That increased the Micro-F1 score of the NYT dataset but decreased other measures. The ``w/o focal loss'' experiment removes the adjustment of focal loss, which seems to have a good effect on the RCV1-V2 dataset. For that dataset, the Micro-F1 is slightly better than the state-of-the-art in Table~\ref{tab:results} but worse than our best-performing model on other measures. 
In general, we conclude that it is helpful for the \mymodel model to pay attention to the difficult samples, but too much attention has a negative impact. Thus, focal loss on batch level is the best choice. 

The ``with labels semantics'' experiment uses the linguistic information in the label names, as done in previous work~\cite{jiang-etal-2022-exploiting}. 
We exploit the label semantics by initializing the decoder with RoBERTa embeddings of the original label text.
For each symbolic label, we feed its associated text into RoBERTa to obtain token embeddings. 
We compute the mean over these token embeddings and use it to initialize the corresponding vector in the decoder's embedding and output layer.  
Exploiting label semantics results in slightly lower scores than our best-performing model. Perhaps the reason is that the decoder is simpler than the encoder and uses a different vocabulary, so using label semantics in the decoder has a negative impact. 

Lastly, we examine the effect of using a different encoder. 
Using BERT instead of RoBERTa results in around 1\% decrease on all measures. We have also evaluated different encoder models like Sentence Transformer and XLNET-base, but RoBERTa performed best. 

\subsection{Error Analysis}
\label{subsec:error_analysis}
This section analyzes the type of errors our \mymodel model produces. The examples presented here are from our best-performing model.

\paragraph{WOS} The hierarchy of this dataset is only two levels, with each sample having exactly two labels, one from each level. In $1,630$ out of $9,397$ test samples, the \mymodel model made one or more mistakes while generating labels. Among them were $852$ cases in which the model mispredicted the child label, but the parent label was correct. For example, on one sample, the model predicted 
\texttt{[[a\_11], <unk>, [a\_79],<unk>]} but the ground truth is \texttt{[[a\_72], <unk>, [a\_79], <unk>]}. 
Looking at the hierarchy, we found that the predicted child label \texttt{[a\_11]} and the corresponding ground truth label \texttt{[a\_72]} share the same parent \texttt{[a\_79]}, 
which explains why the model was able to get the parent right. 
For the remaining $778$ wrong cases, the model mispredicted both labels. 
So, the model always predicts the parent label correctly, but when it fails to predict the correct child label, it builds on that error and mispredicts its parent.
This problem is known as exposure bias~\cite{yang-etal-2018-sgm}.
In other words, the model has learned the dataset's child-to-parent relationships but faces difficulties generating the correct child labels in the first place.

\paragraph{NYT}
Unlike WOS, the NYT dataset has an eight-level hierarchy. 
We observe different types of errors in $3,791$ out of $7,292$ test samples. Among the errors, the model generates shorter sequences for $1,805$ samples and longer sequences for $1,699$ samples, while only $287$ samples have the same number of tokens as the gold standard. This type of error indicates that the model has difficulty generating the end-of-sequence token correctly. Let us further investigate the three cases. For shorter sequence cases, the model fails to predict a label that starts a new path somewhere toward the end of the sequence as in the following ground truth sequence: 
\texttt{[[a\_52], <unk>, [a\_48], [a\_30], <unk>, [a\_47], [a\_29], <unk>, [a\_46], 
[a\_36], [a\_24], <unk>, [a\_23], <unk>]}. In this label sequence, the model fails to predict the label \texttt{[a\_36]}, which comes before the root node with no previous children, but correctly predicts two labels after the first generated label. This phenomenon indicates that the model depends on the sample text at the beginning of the sequence but gradually shifts its focus to depend on the predicted labels as it generates more of them. 

For longer sequences, the model fails to select the correct starting level from the eight hierarchy levels. For example, given the ground truth sequence \texttt{[[a\_99], <unk>, [a\_85] <unk>]}, it predicts \texttt{[[a\_12], <unk>, [a\_2], <unk>, [a\_1][a\_99], <unk>,[a\_0][a\_85], <unk>]}, which includes the ground truth but adds extra labels. This shows that the model captures the hierarchy but generates labels from deeper levels than necessary.

Finally, for sequences with the same length, the model suffers from the exposure bias problem like on the WOS dataset.  

\paragraph{RCV1-V2}
This dataset has a four-level hierarchy.
The model makes different errors generating the labels in $260,415$ out of $781,265$ test samples. 
Among them, it generates shorter sequences for $124,622$ samples and longer sequences for $104,336$ samples, while only $31,457$ samples have the same length as the gold standard. 
Investigating these three types of mistakes, we find similar errors to those of NYT. When comparing the errors with the ones on the NYT dataset, we notice that the model, in general, erroneously produces many shorter sequences than longer ones.

\section{Conclusion and Future Work}
\label{sec:conclusion}
The paper introduces \mymodel, a sequence-to-sequence model using RoBERTa as an encoder and an autoregressive decoder for the HTC task. It does not have an encoder for the label hierarchy but learns it from the label sequences during training. 
The model shows that organizing label sequences from children to parents, rather than the opposite order~\cite{wang-etal-2022-incorporating, jiang-etal-2022-exploiting, zhou-etal-2020-hierarchy, deng2021htcinfomax, chen-etal-2021-hierarchy} (see Table~\ref{tab:model_comparison}), is more effective. 
\mymodel demonstrates that label semantics or encoding the label hierarchy are not necessary for good performance. 
As a result, \mymodel is a flexible model with fewer requirements and easy-to-replace components. 
In future work,
we plan to further investigate the exposure bias problem and also evaluate the model on non-hierarchical multi-label classification tasks. 
%
%
%
\section*{Limitations}
Our \mymodel model is an English model because we have fine-tuned it on three English benchmark datasets. 
Nevertheless, the model is extendable to other languages since the decoder uses symbolic labels and does not utilize label semantics. Also, the encoder could be easily replaced by a multilingual transformer encoder. 

\begin{ack}
This work was co-funded by the DFG as part of the UnknownData Project - Grant No. 460676019.
\end{ack}


\bibliography{references}

\end{document}